\documentclass[conference]{IEEEtran}
\IEEEoverridecommandlockouts
\usepackage{cite}
\usepackage{amsmath,amssymb,amsfonts}
\usepackage{algorithmic}
\usepackage{graphicx}
\usepackage{textcomp}
\usepackage{url}
\usepackage{breakurl}
\usepackage{xcolor}
\usepackage{subcaption}
\usepackage{multirow}
\usepackage{array}
\usepackage{ragged2e}
\usepackage{hyperref}
\usepackage{float}

\def\BibTeX{{\rm B\kern-.05em{\sc i\kern-.025em b}\kern-.08em
    T\kern-.1667em\lower.7ex\hbox{E}\kern-.125emX}}
\begin{document}

\title{Transfer Learning via Lexical Relatedness: \\ A Sarcasm and Hate Speech Case Study\\}

\author{
\IEEEauthorblockN{Angelly Cabrera}
\IEEEauthorblockA{University of Southern California}
\and
\IEEEauthorblockN{Linus Lei} 
\IEEEauthorblockA{University of Southern California}
\and
\IEEEauthorblockN{Antonio Ortega} 
\IEEEauthorblockA{University of Southern California}
}

\maketitle

\begin{abstract}
Detecting hate speech in non-direct forms, such as irony, sarcasm, and innuendos, remains a persistent challenge for social networks. Although sarcasm and hate speech are regarded as distinct expressions, our work explores whether integrating sarcasm as a pre-training step improves implicit hate speech detection and, by extension, explicit hate speech detection. Incorporating samples from ETHOS, Sarcasm on Reddit, and Implicit Hate Corpus, we devised two training strategies to compare the effectiveness of sarcasm pre-training on a CNN+LSTM and BERT+BiLSTM model. The first strategy is a single-step training approach, where a model trained only on sarcasm is then tested on hate speech. The second strategy uses sequential transfer learning to fine-tune models for sarcasm, implicit hate, and explicit hate. Our results show that sarcasm pre-training improved the BERT+BiLSTM’s recall by 9.7\%, AUC by 7.8\%, and F1-score by 6\% on ETHOS. On the Implicit Hate Corpus, precision increased by 7.8\% when tested only on implicit samples. By incorporating sarcasm into the training process, we show that models can more effectively detect both implicit and explicit hate. \textcolor{red}{Note: This paper contains offensive and derogatory language shown only for demonstration.} 
\end{abstract}

\section{Introduction}

A key challenge in specialized machine learning is the lack of sufficient data for a given task. However, it is often possible to identify other datasets that are lexically or semantically related, even if they do not exactly correspond to the task at hand. This paper considers one such scenario, where sarcasm data is used to pre-train for hate speech detection, but we expect similar situations to arise in other applications. 

Hate speech refers to derogatory and discriminatory language based on race, gender, sexual orientation, religion, or nationality. Hate speech is defined as containing any of the following: (1) negative stereotypes, (2) dehumanizing language, and (3) violent expressions~\cite{schafer2024counteract}. Online exposure to hate speech is widespread. A 2017 study found that 53\% of American and 31\% of German respondents reported having encountered hate speech within the last three months~\cite{hawdon2017exposure}. Among young adults aged 18 to 25, exposure rates were especially high at 71\%~\cite{reichelmann2021hate}. However, exposure is not evenly distributed. The LGBTQ+ community is one of the most frequent targets of online hate. Transgender people are more likely to experience depression, anxiety, and hate crimes resulting from online abuse~\cite{stefanita2021hate}. 

While explicit hate speech (EHS), which contains overtly offensive language, is easier to identify, implicit hate speech (IHS) is subtle and context-dependent, often masked by humor or ambiguity. Despite efforts to flag abusive content, much of the focus has been on EHS, while IHS remains understudied and harder to classify~\cite{elsherief-etal-2021-latent}. State-of-the-art (SOTA) models used for hate speech detection, such as deBERTa v3, HateBERT, and BERT, while highly effective for EHS, struggle to classify IHS~\cite{ocampo-etal-2023-depth}. Recent studies on toxic language suggest that sarcasm is a common rhetorical tool for masking abusive intent~\cite{Madhyastha_Founta_Specia_2023}. Sarcasm, like implicit hate, frequently relies on indirect phrasing, misdirection, and tone that contradicts literal meaning. Both forms may use innocuous language that conceals negative sentiment or aggression, making them difficult for models to detect. These parallels indicate a potential overlap not only in how these forms are expressed but also in the types of linguistic cues they employ. To illustrate this, \autoref{tab:sarcasm_vs_implicit_hate} presents a side-by-side comparison of a sarcastic comment mocking segregationist logic and an implicit hate comment expressing hostility toward migrant and Muslim communities.

Our main hypothesis is that using a lexically or semantically related task can compensate for the lack of data for the target task. For any category with scarce explicit labels, a related category can be identified based on measures such as shared vocabulary, distributional similarity in embeddings, or thematic overlap, without being an exact synonym. Training models on sarcasm, for example, can improve their ability to detect implicit hate speech by increasing the amount of training data available. By extension, this method would also enhance performance on explicit hate. \textit{We propose that pre-training on sarcasm helps models recognize indirect hostile expressions and rhetorical misdirection, capabilities that are crucial for identifying covert hate}. In this work, we explore and evaluate the benefits of incorporating sarcasm as an auxiliary task in the hate speech detection pipeline. 

\begin{table}[htbp]
    \centering
    \captionsetup{justification=centering}
    \small
    \renewcommand{\arraystretch}{1.3} % Row height
    \begin{tabular}{p{2cm}|p{6cm}} 
        \textbf{Category}     & \textbf{Example}        
        \\ \hline
        \textbf{Sarcasm} & ``It's not segregation if it's done for the `right' reason, and I know what the `right' reason is because I'm right.'' \\  
        \textbf{Implicit Hate} & ``Well im glad that I live in Serbia  migrants and islamists are not welcome here!'' \\  
    \end{tabular}
    \caption{Examples of sarcasm~\cite{SARC} and implicit hate speech~\cite{mollas_ethos_2022}.}
    \label{tab:sarcasm_vs_implicit_hate}
    \vspace{-3mm}
\end{table}

\section{Related Work}
\label{sec:related_work}
Detecting IHS is a persistent challenge in natural language processing due to its reliance on indirect cues, context, and rhetorical devices. While transformer-based models like deBERTa v3 and HateBERT perform well on explicit hate detection, their effectiveness drops significantly on implicit hate due to being largely optimized for overt abusive language\cite{ocampo-etal-2023-depth}. Studies such as that of ElSherief et al. introduced the Implicit Hate Corpus to benchmark models against more subtle forms of abuse, also revealing how traditional classifiers often miss hostile covert language that avoids slurs\cite{elsherief-etal-2021-latent}.

To address these limitations, researchers have explored alternatives beyond architectural improvements. One avenue involves augmenting datasets with synthetic samples using generative models. For instance, Hartvigsen et al.~introduced ToxiGen, a dataset designed to improve IHS detection with synthetically generated hate speech samples\cite{hartvigsen2022toxigenlargescalemachinegenerateddataset}. Another example is Almohaimeed et al., who used GPT-4 to categorize datasets into normal, explicit, and IHS labels, which were then validated by human reviewers\cite{almohaimeed2024implicit}. However, this method proved to have its own challenges, as GPT-4 misclassified statements when the target was an individual, struggled to differentiate between opinion and sarcasm, and exhibited bias against certain demographic groups. For example, the model incorrectly classified the statement \textit{``Women make much better music than most men''} as hateful for making an assertion of superiority. When compared to human annotations, GPT-4 recorded low precision (33.1\%) and recall (45.5\%). 

Existing work related to sarcasm has sought to break down its vocabulary so that models are able to detect and replicate it. For example, Mishra et al. introduced a framework for sarcasm generation that involves providing a literal negative opinion and translating it into sarcasm. Their work proposes that models trained to detect sarcasm may become more sensitive to subtle hostility \cite{mishra-etal-2019-modular}. Sarcasm and hate speech detection are typically treated independently, but studies show that sarcasm and IHS both use figurative speech. Waseem et al.~found that IHS does not directly indicate abuse \cite{waseem-etal-2017-understanding}. 
Its true nature is often obscured by the use of ambiguous terms, figurative language, and a lack of profanity and violent rhetoric. Notable examples of figurative speech used to mask abusive intent include humor, hyperboles, metonymy, and sarcasm.

Other works on sarcasm and hate speech detection, such as \cite{khan2022bichat,SaviniEdoardo,Bhardwaj2022}, use transfer learning by first training their models on one or more related datasets before fine-tuning them on their target task. For instance, Savini et al. combined tweets and IMDb movie reviews to expose their BERT model to abstract, polarized, and figurative speech, then fine-tuned it for sarcasm detection \cite{SaviniEdoardo}. Khan et al.\cite{khan2022bichat} and Bhardwaj et al.\cite{Bhardwaj2022} followed a similar process for their respective tasks, but each focused exclusively on either sarcasm detection or hate speech detection. In all these cases, the intermediate task is aimed at improving the same end task, rather than a different one. In other words, they do not train on sarcasm with the goal of improving hate speech detection, or vice versa. \textit{Our approach differs in that we explicitly use sarcasm as the intermediate task within the transfer learning process, with the goal of improving hate speech detection}. Rather than relying solely on data augmentation or adversarial training, we treat sarcasm as a proxy task that trains models to identify implicit language, preparing them to better recognize covert hate.

\section{Proposed Methodology}
Our methodology consists of three stages. First, for a target task with limited labeled data, such as IHS detection, we identify one or more related source tasks that are lexically or semantically similar. In our case, sarcasm was chosen as the related task. Next, we measure the degree of relatedness between the source and target tasks using surface-level lexical similarity metrics. Finally, we integrate the selected related task into the pipeline as an intermediate step in a transfer learning framework and evaluate two training strategies: (1) single-step training on the related task before evaluating on the target task, and (2) sequential fine-tuning from the related task to the target task. We then compare different model architectures under these strategies to assess how pre-training on the related task influences the target task performance.

\subsection{Data \& Similarity Measures}
Discussed in \autoref{sec:related_work}, prior work shows that sarcasm and implicit hate use figurative speech to mask true intent. Building on these observations, we examine whether their vocabularies overlap at the surface level. We measure overlap between sarcasm, IHS, and EHS using Jaccard similarity with $n$-grams and Jensen-Shannon divergence (JSD). These statement-level measures capture surface-level patterns that are not obvious in semantic embeddings alone. We, therefore, prioritize them over string-based methods such as longest common subsequence (LCS) and Levenshtein distance.

Jaccard similarity measures overlap by computing the intersection-over-union of two word sets, which makes it well-suited for cross-corpus comparison between our datasets \cite{pradhan2015review}. We extend this by computing Jaccard similarity over the top 1,000 unigrams and bigrams extracted from randomly sampled subsets of each dataset, allowing us to retain some contextual information. Jensen-Shannon divergence, in contrast, captures distributional similarity by accounting for word frequency. A lower JSD score indicates that two corpora use similar vocabulary with similar frequency patterns~\cite{Jiapeng2020}.

We perform these across three datasets: Sarcasm on Reddit~\cite{SARC}, ETHOS~\cite{mollas_ethos_2022}, and the Implicit Hate Corpus~\cite{elsherief-etal-2021-latent}. Summarized in \autoref{tab:dataset_info}, these datasets provide coverage of explicit, implicit, and non-literal rhetorical expressions. \autoref{tab:explicit_vs_implicit_hate} shows a side-by-side comparsion between explicit hate (``Go back to Mexico, racist!'') and implicit hate (``If it's white, it's right''). The implicit hate example resembles the sarcasm example previously shown in \autoref{tab:sarcasm_vs_implicit_hate}. Previous work has used the Implicit Hate Corpus for its clear labeling of implicit and explicit hate speech \cite{elsherief-etal-2021-latent}. However, the version used in this study, which includes the original Twitter-sourced data, is imbalanced, as implicit and explicit samples make up only 33\% of the dataset. Therefore, we applied class weighting during training. ETHOS, which includes a mix of explicit and implicit hate, uses continuous labels ranging from 0 to 1. Based on a qualitative dataset evaluation, we binarized these labels by treating scores $\geq$ 0.33 as hate speech. Additionally, Sarcasm on Reddit samples were filtered based on vote patterns, retaining only those with over 10 upvotes and no downvotes to improve the representation of ambiguous, non-hateful language.

\begin{table}[htbp]
\centering
\captionsetup{justification=centering}
\small
\renewcommand{\arraystretch}{1.3} 
\setlength{\tabcolsep}{10pt} 
\begin{tabular}{c|c|c|c}
    \textbf{Dataset} & \textbf{Source} & \textbf{Label Type} & \textbf{Size} \\
    \hline
    Sarcasm on Reddit & Reddit & Binary & 104,161 \\
    Implicit Hate Corpus & Twitter & Multi-label & 21,480 \\
    ETHOS & Mixed & Continuous & 998 \\
\end{tabular}
\caption{Overview of datasets used in this study.}
\label{tab:dataset_info}
\vspace{-2mm}
\end{table}

\begin{table}[htbp]
    \centering
    \captionsetup{justification=centering}
    \small
    \renewcommand{\arraystretch}{1.3}
    \setlength{\tabcolsep}{15pt} 
    \begin{tabular}{c|c} 
        \textbf{Category}     & \textbf{Example}        
        \\ \hline
        \textbf{Explicit Hate} & ``Go back to Mexico, racist!'' \\  
        \textbf{Implicit Hate} & ``If it's white, it's right'' \\  
    \end{tabular}
    \caption{EHS and IHS sample comparison~\cite{mollas_ethos_2022}.}
    \label{tab:explicit_vs_implicit_hate}
    \vspace{-3mm}
\end{table}

\subsection{Training Strategies}
\begin{figure}[htbp]
    \centering
    \captionsetup{justification=centering}
    \includegraphics[width=\linewidth, keepaspectratio]{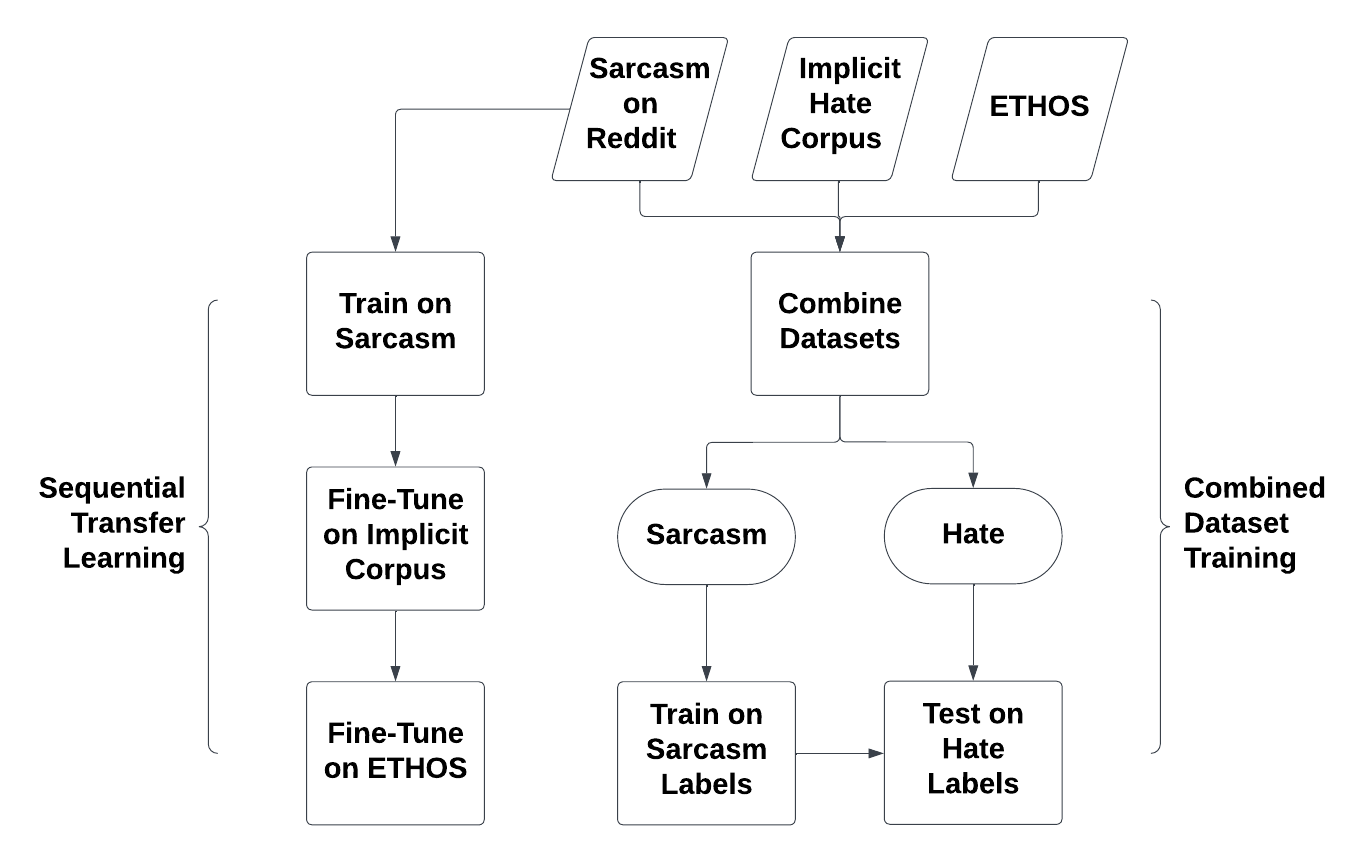}
    \caption{Training strategy flowchart.}
    \vspace{-3mm}
    \label{fig:training_flowchart}
\end{figure}

We integrated sarcasm pre-training into our transfer learning framework using the two strategies described next and shown in \autoref{fig:training_flowchart}. 

\textbf{Single-Step Training Using Combined Dataset}:
In the single-step approach, we group the 4 labels (neutral, sarcasm, IHS, EHS) into two classes (sarcasm vs non-sarcasm) as shown in \autoref{tab:class_labels}. We first train the model to separate sarcasm from non-sarcasm and then evaluate its ability to separate IHS from EHS without using hate-specific fine-tuning. 
The dataset is split once into training and test sets. We first train and evaluate on sarcasm labels, then take the same test set, replace it with the hate speech labels, and evaluate again. This setup assesses whether sarcasm training alone provides transferable information for hate speech detection. 
At first glance, the metrics shown in \autoref{tab:model_metrics} indicate that this method increased the BERT+BiLSTM's and CNN+LSTM's precision by roughly 30\%, achieving 92\% and 85\% precision, respectively. Note that this approach cannot be viewed as a standard example of transfer learning. This is because 
IHS and EHS examples were used during preprocessing, along with other data in different categories (neutral and sarcasm), but without their true labels (only sarcasm vs non-sarcasm). 

\begin{table}[htbp]
    \centering
    \captionsetup{justification=centering}
    \small
    \renewcommand{\arraystretch}{1.3} % Row height
    \setlength{\tabcolsep}{15pt} % Column spacing
    \begin{tabular}{c|c|c} 
        \textbf{Class Label} & \textbf{\textit{Sarcasm}} & \textbf{\textit{Hate}} \\
        \hline
        Neutral & 0 & 0 \\
        Sarcasm & 1 & 0 \\
        Implicit Hate & 1 & 1 \\
        Explicit Hate & 0 & 1 \\
    \end{tabular}
    \caption{Sarcasm and hate labels binarized by class.}
    \label{tab:class_labels}
\end{table}

\textbf{Sequential Transfer Learning Approach}:
Alternatively,  we train the models sequentially and transfer the learned weights between tasks. This method allows the models to progressively adapt to sarcasm, implicit hate, and explicit hate detection tasks. Hence, we first train our model to identify implicit language patterns. For the sarcasm detection task, we trained the models using both the Reddit comment and its corresponding parent comment as separate input streams to provide additional context. \autoref{tab:dataset_examples} shows an example of the difficulty progression that the models are seeing. 

Savini et al.~used transfer learning by fine-tuning a BERT-based model on a series of intermediate tasks, from tweets to movie reviews, before applying it to the target task of sarcasm detection by training the model on datasets that contained progressively more examples of abstract, polarized, and figurative speech~\cite{SaviniEdoardo}. This approach led to an overall improvement in F1-scores with some minor fluctuation between intermediate tasks. Mozafari et al.~also applied transfer learning for hate speech detection by fine-tuning a pre-trained BERT model on downstream tasks using progressively more complex architectures, using 
English Wikipedia and BookCorpus for training and upgrading their model architecture by inserting a Bi-LSTM layer and finally adding a CNN layer~\cite{mozafari2019bertbasedtransferlearningapproach}.

Given the results of the sequential transfer learning approach, which we explore more in \autoref{sec:results}, it will become the primary focus of this paper. 

\begin{table}[htbp]
    \centering
    \captionsetup{justification=centering}
    \small
    \renewcommand{\arraystretch}{1.3} % Row height
    \label{tab:dataset_examples}
    \begin{tabular}{p{3cm}|p{5cm}}
        \textbf{Dataset} & \textbf{Example} 
        \\ \hline
        \textbf{Sarcasm on Reddit} & ``THEY PAY IN EUROS SO IT'S A SOCIALIST COUNTRY'' 
        \\ \hline
        \textbf{Implicit Hate Corpus} & ``deportable illegals? what? every illegal alien parasite in the usa are deportable illegal aliens.'' 
        \\ \hline
        \textbf{ETHOS} & ``You should know women's sports are a joke.'' 
    \end{tabular}
    \caption{Sample training data (positive samples).}
    \vspace{-5mm}
    \label{tab:dataset_examples}
\end{table}

\subsection{Models}
\begin{figure*}[htbp] 
    \centering
    \captionsetup{justification=centering}
    % First image
    \begin{subfigure}{\textwidth}
        \centering
        \includegraphics[width=0.85\textwidth]{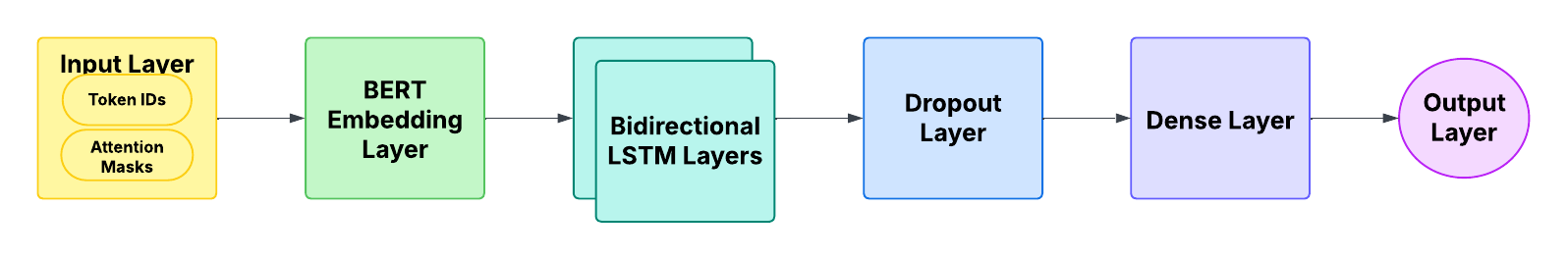} 
        \caption{BERT+BiLSTM model architecture.}
        \label{fig:bert_discriminator_model}
    \end{subfigure}
    \vspace{1em} % Space between the images
    % Second image
    \begin{subfigure}{\textwidth}
        \centering
        \includegraphics[width=1.05\textwidth]{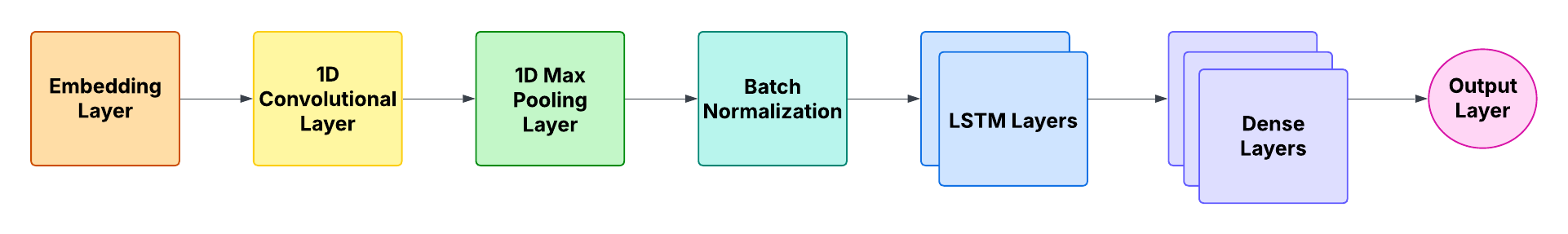} 
        \caption{CNN+LSTM model architecture.}
        \vspace{-5mm}
        \label{fig:conv_discriminator_model}
    \end{subfigure}
    \label{fig:stacked_models}
    \caption{Discriminator model architectures.}
\end{figure*}

Long Short-Term Memory (LSTM) networks, especially bidirectional LSTMs (BiLSTMs), excel at processing sequential data because they capture context from preceding and succeeding words \cite{KrisdiantoRicky, rawat2024hate, khan2022bichat}. Transformer-based models such as BERT have likewise achieved strong results in sentiment analysis and are especially effective for tasks involving nuanced language like sarcasm \cite{Bhardwaj2022}. Convolutional Neural Networks (CNNs) are also useful in text classification thanks to their ability to extract local and spatial features, and prior work has shown that combining CNNs with transformers or LSTMs can improve hate speech detection \cite{khan2022bichat}. Based on these findings, we designed a BERT+BiLSTM architecture (\autoref{fig:bert_discriminator_model}) using BERT as the embedding layer, followed by two BiLSTM layers. We incorporated a dropout layer (20\%) before the dense layers and set the learning rate to $5e^{-5}$ to balance training time with performance. 

Initially, the embedding layer was made trainable, but it resulted in the BERT+BiLSTM having $\approx 67$ million parameters and longer training times. We also found that setting this layer as trainable did not improve its performance. We reduced the model size to $\approx 470,000$ parameters and experimented with $\ell_1$ regularization to penalize large weights. Given that CNNs and LSTMs also work well for this area of research, we also developed an alternative model (\autoref{fig:conv_discriminator_model}) with convolutional and LSTM layers for benchmark comparison. This model was significantly smaller than the BERT+BiLSTM (70,000 parameters). We applied batch normalization to the CNN outputs before feeding them into the LSTM layers to experiment with different regularization techniques. 

Model performance was primarily evaluated using precision, recall, and F1-score to measure their ability to detect hate speech while minimizing false positives. We also included Area Under the Curve (AUC) to measure the model’s discrimination ability across different confidence thresholds, and Matthews Correlation Coefficient (MCC) as a balanced performance measure for class imbalances.

\section{Results}
\label{sec:results}
\subsection{Pre-training Results}
\autoref{tab:model_metrics} shows the performance of the BERT+BiLSTM and CNN+LSTM using the single-step training approach. Although the results for hate speech are promising (92\% and 85\% precision, respectively), the models significantly underperformed when tested on a separate dataset. Given that IHS and EHS labels were merged during this approach, we cannot attribute performance gains directly to sarcasm pre-training, as the labels limit its interpretability.

\autoref{tab:CNN_metrics_v2} and \autoref{tab:BERT_metrics_v2} present the results of the sequential transfer learning approach. Between the two models, the BERT+BiLSTM showed the most consistent improvement across tasks, achieving over 70\% in all metrics for sarcasm detection and having the strongest performance on ETHOS. It also maintained moderate-to-high recall on both the Implicit Hate Corpus and ETHOS. The steady increase in AUC values across tasks suggests better discrimination between hateful and non-hateful content, and its MCC score reflects a moderate correlation between predictions and true labels. 
In contrast, \autoref{tab:CNN_metrics_v2} shows that the CNN+LSTM struggled to generalize to unseen data. This contrast highlights the sensitivity of CNNs to dataset size and variability. Unlike transformer-based architectures, CNNs rely primarily on local lexical and structural cues present in the training data, limiting their ability to generalize across heterogeneous corpora \cite{khan2022bichat}.

We directly assessed the effect of sarcasm pre-training by comparing BERT+BiLSTM’s performance on the Implicit Hate Corpus with and without pre-training. In this setup, the model classified IHS and EHS vs. non-hateful samples. As shown in \autoref{tab:implicit_sarcasm_pre-training}, sarcasm pre-training increased precision by 4.4\% and AUC by 1.9\%. We ran a similar evaluation on ETHOS, using 40\% of the data for training and 60\% for testing. As shown in \autoref{tab:ETHOS_sarcasm_pre-training}, sarcasm pre-training boosted recall by 9.7\%, F1-score by 6\%, and improved MCC and AUC by 7.9\% and 7.8\%, respectively. In practical terms, this means the model captured more abusive content while reducing false positives. Obtaining this balance is imperative for moderation systems, since reducing false positives maintains the integrity of user interactions while still ensuring that harmful content is identified and handled accordingly.

\begin{table*}[htbp]
    \centering
    \captionsetup{justification=centering}
    \small
    \renewcommand{\arraystretch}{1.3} % Row height
    \setlength{\tabcolsep}{12pt} % Column spacing
    \begin{tabular}{c|c|c|c|c}
        \textbf{Model} & \textbf{Label} & \textbf{Precision} & \textbf{Recall} & \textbf{F1-Score} 
        \\ \hline
        \multirow{2}{*}{BERT+BiLSTM}
        & Sarcasm & 0.60 & 0.64 & 0.62 
        \\ \cline{2-5}
        & Hate & 0.92 & 0.81 & 0.86 
        \\ \hline
        \multirow{2}{*}{CNN+LSTM} 
        & Sarcasm & 0.52 & 0.59 & 0.55 
        \\ \cline{2-5}
        & Hate    & 0.85 & 0.81 & 0.83 
        \\ 
    \end{tabular}
    \caption{BERT+BiLSTM and CNN+LSTM performance using single-step approach.}
    \label{tab:model_metrics}
\end{table*}

\begin{table*}[htbp]
    \centering
    \captionsetup{justification=centering}
    \small
    \renewcommand{\arraystretch}{1.3} 
    \setlength{\tabcolsep}{12pt} 
    \begin{tabular}{c|c|c|c|c|c}
        \textbf{Dataset} & \textbf{Label} & \textbf{Precision} & \textbf{Recall} & \textbf{F1-Score} & \textbf{AUC} 
        \\ \hline
        \textbf{Sarcasm on Reddit} 
        & Sarcasm & 0.65 & 0.60 & 0.63 & 0.65 
        \\ \hline
        \multirow{2}{*}{\textbf{Implicit Hate Corpus}} 
        & Implicit Hate & 0.51  & 0.45 & 0.48 & 0.66 
        \\ \cline{2-6}
        & Hate & \textbf{0.93} & \textbf{0.78} & \textbf{0.85}  & \textbf{0.66} 
        \\ \hline
        \textbf{ETHOS} & Hate & 0.58 & 0.59  & 0.59  & 0.51 
        \\ \hline
    \end{tabular}
    \caption{CNN+LSTM intermediate and target task performance using sequential approach.}
    \label{tab:CNN_metrics_v2}
\end{table*}

\begin{table*}[htbp]
    \centering
    \captionsetup{justification=centering}
    \small % Keep the table compact
    \renewcommand{\arraystretch}{1.3} 
    \setlength{\tabcolsep}{12pt} 
    \begin{tabular}{c|c|c|c|c|c}
        \textbf{Dataset} & \textbf{Label} & \textbf{Precision} & \textbf{Recall} & \textbf{F1-Score} & \textbf{AUC} 
        \\ \hline
        \textbf{Sarcasm on Reddit} & Sarcasm & 0.73 & 0.75 & 0.73 & 0.78 
        \\ \hline
        \multirow{2}{*}{\textbf{Implicit Hate Corpus}} 
        & Implicit Hate & 0.60  & 0.67 & 0.60 & 0.70 
        \\ \cline{2-6}
        & Hate & 0.67 & 0.73  & 0.70  & 0.75 
        \\ \hline
        \textbf{ETHOS} & Hate & \textbf{0.77} & \textbf{0.87} & \textbf{0.82} & \textbf{0.81} \\ \hline
    \end{tabular}
    \caption{BERT+BiLSTM intermediate and target task performance using sequential approach.}
    \label{tab:BERT_metrics_v2}
\end{table*}

\begin{table}[htbp]
    \centering
    \captionsetup{justification=centering}
    \small
    \renewcommand{\arraystretch}{1.3} % Row height
    \begin{tabular}{c|c|c|c}
        \textbf{Metric} & \textbf{\shortstack{Implicit Corpus \\ (No Sarcasm \\ Pre-training)}} & \textbf{\shortstack{Implicit Corpus \\ (Sarcasm \\ Pre-training)}} & \textbf{Change} 
        \\ \hline
        Precision & 0.626 & 0.670 & +4.4\% 
        \\ \hline
        Recall    & 0.726 & 0.730 & +0.4\% 
        \\ \hline
        F1-Score  & 0.672 & 0.699 & +2.7\% 
        \\ \hline
        MCC       & 0.452 & 0.460 & +0.8\% 
        \\ \hline
        AUC       & 0.731 & 0.750 & +1.9\% \\
    \end{tabular}
    \caption{BERT+BiLSTM Implicit Hate Corpus performance with and without sarcasm pre-training.}
    %\vspace{-2.5mm}
    \label{tab:implicit_sarcasm_pre-training}
\end{table}

\begin{table}[htbp]
    \centering
    \captionsetup{justification=centering}
    \small
    \renewcommand{\arraystretch}{1.3} % Row height
    \begin{tabular}{c|c|c|c}
        \textbf{Metric} & \textbf{\shortstack{ETHOS \\ (No Sarcasm \\ Pre-training)}} & \textbf{\shortstack{ETHOS \\ (Sarcasm \\ Pre-training)}} & \textbf{Change} 
        \\ \hline
        Precision & 0.734 & 0.771 & +3.7\%
        \\ \hline
        Recall    & 0.769 & 0.866 & +9.7\%
        \\ \hline
        F1-Score  & 0.756 & 0.816 & +6.0\%
        \\ \hline
        MCC       & 0.465 & 0.544 & +7.9\%
        \\ \hline
        AUC       & 0.732 & 0.810 & +7.8\% \\
    \end{tabular}
    \caption{BERT+BiLSTM ETHOS performance with and without sarcasm pre-training.}
    \label{tab:ETHOS_sarcasm_pre-training}
    %\vspace{-3mm}
\end{table}

We further isolated the impact of sarcasm pre-training by testing the BERT+BiLSTM on implicit-only samples from the Implicit Hate Corpus. Results shown in \autoref{tab:implicit_labels_sarcasm_pre-training} show that the BERT+BiLSTM performed lower on implicit samples than on combined IHS and EHS samples. Nevertheless, sarcasm pre-training still raised precision by 8.8\%. On the other hand, the 7.4\% drop in recall indicates that the model also missed more implicit samples, likely a result of the model becoming more conservative in its classifications. \autoref{tab:corpus_sarcasm_pre-training} compares performance on implicit-only vs. combined IHS and EHS samples only when sarcasm pre-training is applied. Here, overall performance improved by 7.8\%, with gains of 6.9\% in recall, 10.3\% in F1-score, and 5.1\% in AUC. These results confirm that sarcasm pre-training strengthens the model’s ability to recognize implicit hate speech and better separate hateful from non-hateful content. 

Finally, \autoref{tab:corpus_NO_sarcasm_pre-training} shows the same comparison without sarcasm pre-training. In this case, the BERT+BiLSTM still improved, though the gains were smaller and less balanced. The F1-score increased by 7.6\%, compared to a 10.3\% gain with sarcasm pre-training, and AUC rose by only 3.7\%. This suggests that sarcasm pre-training not only boosts overall performance but also helps achieve a better balance between minimizing false positives and false negatives.

While prior work has applied transfer learning separately to sarcasm detection and to hate speech detection, the source and target tasks in those setups are the same (e.g., pre-training on one sarcasm dataset, then fine-tuning on another). None have used sarcasm detection as the source task for improving hate speech detection. In contrast, our work uses transfer learning, but as a cross-task transfer where we first train our models on sarcasm, then fine-tune on hate speech, with the goal of enhancing hate speech detection. For reference, Savini et al. \cite{SaviniEdoardo} evaluated transfer learning on two sarcasm datasets, Sarcasm V2~\cite{oraby-etal-2016-creating} and SARC \cite{khodak-etal-2018-large}. Mozafari et al. \cite{mozafari2019bertbasedtransferlearningapproach} evaluated their model on two datasets created by Waseem (now published under the name Talat) et al. \cite{Talat2018} and David et al. \cite{davidson-etal-2019-racial}. \autoref{tab:metric_comparisons} compares F1-scores between our training approach with the aforementioned approaches. Although these results are not directly comparable, they help contextualize the strength of transfer learning to improve hate speech detection. They also show that sarcasm pre-training can offset limited training data for the target task, achieving results comparable to \cite{SaviniEdoardo}, despite using fewer training samples.

\begin{table}[htbp]
    \centering
    \captionsetup{justification=centering}
    \small
    \renewcommand{\arraystretch}{1.3} % Row height
    \begin{tabular}{c|c|c|c}
        \textbf{Metric} & \textbf{\shortstack{Implicit Labels \\ Only \\ (No Sarcasm \\ Pre-training)}} & \textbf{\shortstack{Implicit Labels \\ Only \\ (Sarcasm \\ Pre-training)}} & \textbf{Change} 
        \\ \hline
        Precision & 0.504 & 0.592 & +8.8\%
        \\ \hline
        Recall    & 0.729 & 0.661 & -7.4\% 
        \\ \hline
        F1-Score  & 0.596 & 0.596 & +0\% 
        \\ \hline
        MCC       & 0.364 & 0.397 & +3.3\% 
        \\ \hline
        AUC       & 0.694 & 0.699 & +0.5\% \\
    \end{tabular}
    \caption{BERT+BiLSTM Implicit Hate Corpus performance using IHS samples only.}
    \vspace{+3mm}
    \label{tab:implicit_labels_sarcasm_pre-training}
\end{table}

\begin{table}[htbp]
    \centering
    \captionsetup{justification=centering}
    \small
    \renewcommand{\arraystretch}{1.3} % Row height
    \begin{tabular}{c|c|c|c}
        \textbf{Metric} & \textbf{\shortstack{Implicit Labels \\ Only \\ (Sarcasm \\ Pre-training)}} & \textbf{\shortstack{All Hate \\ Labels \\ (Sarcasm \\ Pre-training)}} & \textbf{Change} 
        \\ \hline
        Precision & 0.592 & 0.670 & +7.8\%
        \\ \hline
        Recall    & 0.661 & 0.730 & +6.9\% 
        \\ \hline
        F1-Score  & 0.596 & 0.699 & +10.3\% 
        \\ \hline
        MCC       & 0.397 & 0.460 & +6.3\% 
        \\ \hline
        AUC       & 0.699 & 0.750 & +5.1\% \\
    \end{tabular}
    \caption{BERT+BiLSTM’s performance on IHS-only samples vs. IHS + EHS with sarcasm pre-training.}
    \vspace{+2mm}
    \label{tab:corpus_sarcasm_pre-training}
\end{table}

\begin{table}[H]
    \centering
    \captionsetup{justification=centering}
    \small
    \renewcommand{\arraystretch}{1.3} % Row height
    \begin{tabular}{c|c|c|c}
        \textbf{Metric} & \textbf{\shortstack{Implicit Labels \\ Only \\ (No Sarcasm \\ Pre-training)}} & \textbf{\shortstack{All Hate \\ Labels \\ (No Sarcasm \\ Pre-training)}} & \textbf{Change} 
        \\ \hline
        Precision & 0.504 & 0.626 & +12.2\%
        \\ \hline
        Recall    & 0.729 & 0.726 & -0.3\% 
        \\ \hline
        F1-Score  & 0.596 & 0.672 & +7.6\% 
        \\ \hline
        MCC       & 0.364 & 0.452 & +8.8\% 
        \\ \hline
        AUC       & 0.694 & 0.731 & +3.7\% \\
    \end{tabular}
    \caption{BERT+BiLSTM’s performance on IHS-only samples vs. IHS + EHS without sarcasm pre-training.}
    \label{tab:corpus_NO_sarcasm_pre-training}
\end{table}

\begin{table*}[htbp]
\centering
\captionsetup{justification=centering}
\small
\renewcommand{\arraystretch}{1.5}
\setlength{\tabcolsep}{10pt}
\begin{tabular}{c|c|c|c|c}
    \textbf{Model} & \textbf{Detection Task} & \textbf{Total Sample Size} & \textbf{Test Dataset} & \textbf{F1-Score}
    \\ \hline
    \shortstack{\rule{0pt}{1.3em}BERT+BiLSTM \\ (Sarcasm Pre-training)} & Hate Speech & 126,639 & ETHOS & 81.6
    \\ \hline
    \shortstack{\rule{0pt}{1.3em}BERT+BiLSTM \\ (No Sarcasm Pre-training)}  & Hate Speech & 998 & ETHOS & 75.6
    \\ \hline
    BERT+TransferMDB~\cite{SaviniEdoardo} & Sarcasm & 150,000 & Sarcasm V2 & 80.85 \\
    \hline
    BERT+TransferEmoNetSent~\cite{SaviniEdoardo} & Sarcasm & 150,000 & SARC & 77.53
    \\ \hline
    BERT+LSTM~\cite{mozafari2019bertbasedtransferlearningapproach} & Hate Speech & 3.3B tokens\textsuperscript{*} & Waseem & 86.0
    \\ \hline
    BERT+LSTM~\cite{mozafari2019bertbasedtransferlearningapproach} & Hate Speech & 3.3B tokens\textsuperscript{*} & Davidson & 92.0
    \end{tabular}
    \begin{minipage}{0.80\textwidth}
        \small
        \center{\textsuperscript{*}Token count reported, not number of samples.}
    \end{minipage}
    \caption{Comparison of our training approach with related works.}
    \label{tab:metric_comparisons}
\end{table*}

\subsection{Quantifying Lexical \& Semantic Similarity}
\begin{figure}[htbp]
    \vspace{-5mm}
    \centering
    \captionsetup{justification=centering}
    \includegraphics[width=\linewidth, keepaspectratio]{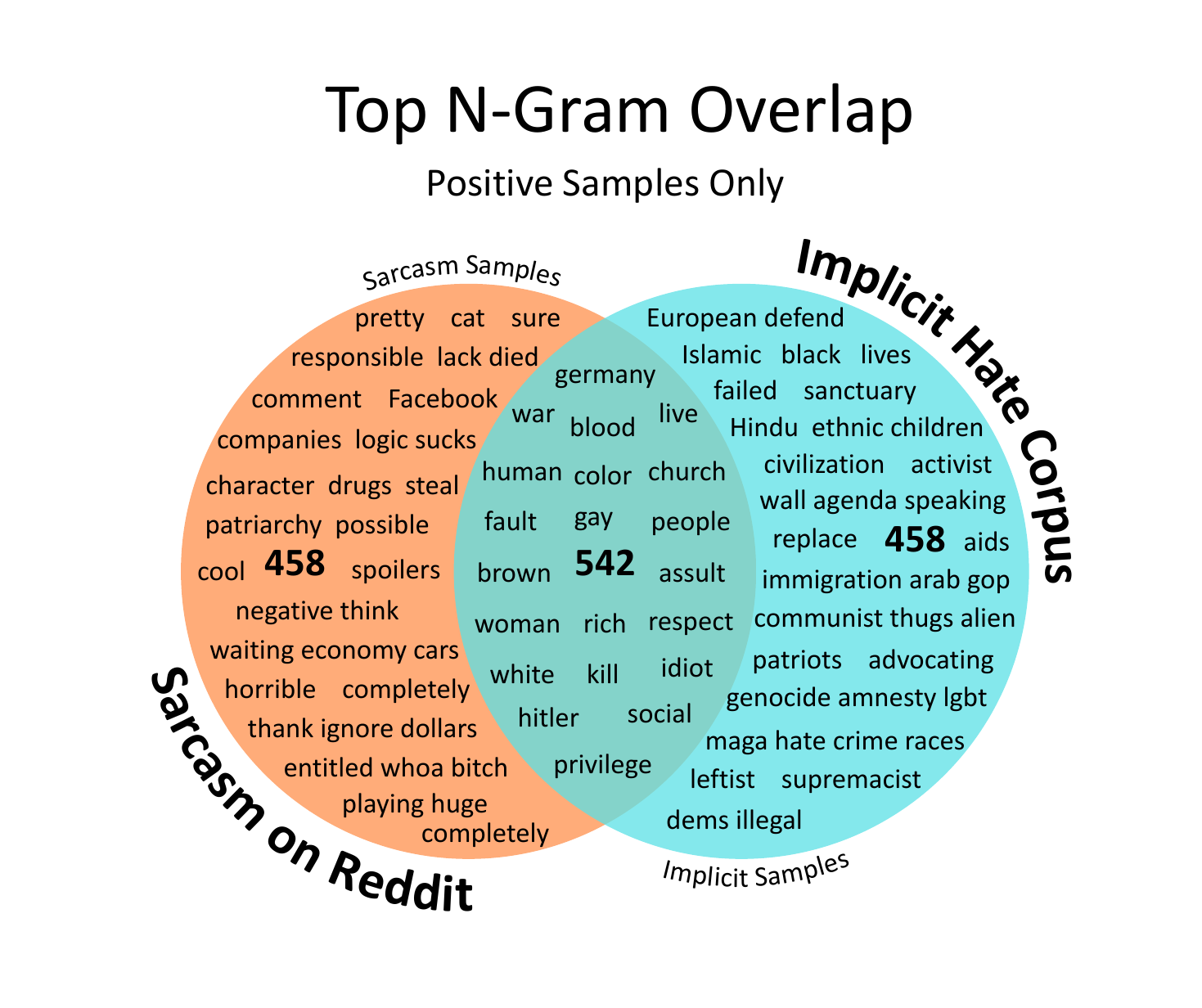}
    \caption{Sarcasm on Reddit vs. Implicit Hate Corpus Venn diagram.}
    \label{fig:jaccard_venn_diagram}
\end{figure}

Surface-level similarity between sarcasm and hate speech was assessed using Jaccard similarity over each dataset’s top-$n$ unigrams and bigrams. We performed 1,000 bootstrapped iterations on randomly sampled subsets, comparing Sarcasm on Reddit with (1) ETHOS and (2) the Implicit Hate Corpus. For these comparisons, we used 5,000 samples for Sarcasm vs. Implicit Hate Corpus and 500 samples for Sarcasm vs.~ETHOS. As shown in \autoref{tab:Jaccard_table}, both comparisons yielded an average Jaccard index of 0.353, suggesting a moderate level of lexical overlap. The Sarcasm vs. ETHOS pair produced the highest individual similarity score (0.576), whereas the Sarcasm vs. Implicit Hate Corpus comparison demonstrated more consistent scores overall. A closer look at the shared $n$-grams reveals which words commonly appear across datasets. \autoref{fig:jaccard_venn_diagram} visualizes one iteration where 54.2\% of the top 1,000 $n$-grams were shared between Sarcasm on Reddit and the Implicit Hate Corpus, with 458 $n$-grams unique to either.

We also computed the JSD between our sarcasm and hate speech datasets. JSD helps us understand not just which words overlap, but how often they appear and how language is distributed across samples. As shown in \autoref{tab:Jensen_Shannon_table}, sarcasm is more closely aligned with the Implicit Hate Corpus (JSD = 0.517) than with ETHOS (JSD = 0.611), which contained mixed IHS and EHS samples. These results align with our Jaccard findings and support our hypothesis that sarcasm may help improve IHS detection based on shared surface-level features, despite differences in intent.

To establish a baseline, we examined Sarcasm on Reddit vs. Sarcasm V2~\cite{oraby-etal-2016-creating}, used to evaluate~\cite{SaviniEdoardo}, to create a baseline for comparison between Sarcasm vs. Implicit Hate Corpus and Sarcasm vs. ETHOS. Shown in \autoref{tab:sarcasm_similarity_metrics}, this comparison yielded a Jaccard score of 0.369 and a JSD of 0.496, both slightly higher than the values for Sarcasm vs. Implicit Hate Corpus. While the absence of significantly higher similarity scores may be attributed to the larger size of Sarcasm on Reddit and the context-dependent, variable nature of sarcasm, the results nonetheless reinforce the validity of our findings.

\begin{figure}[ht]
    \vspace{-5mm}
    \centering
    \captionsetup{justification=centering}
    \includegraphics[width=\linewidth, keepaspectratio]{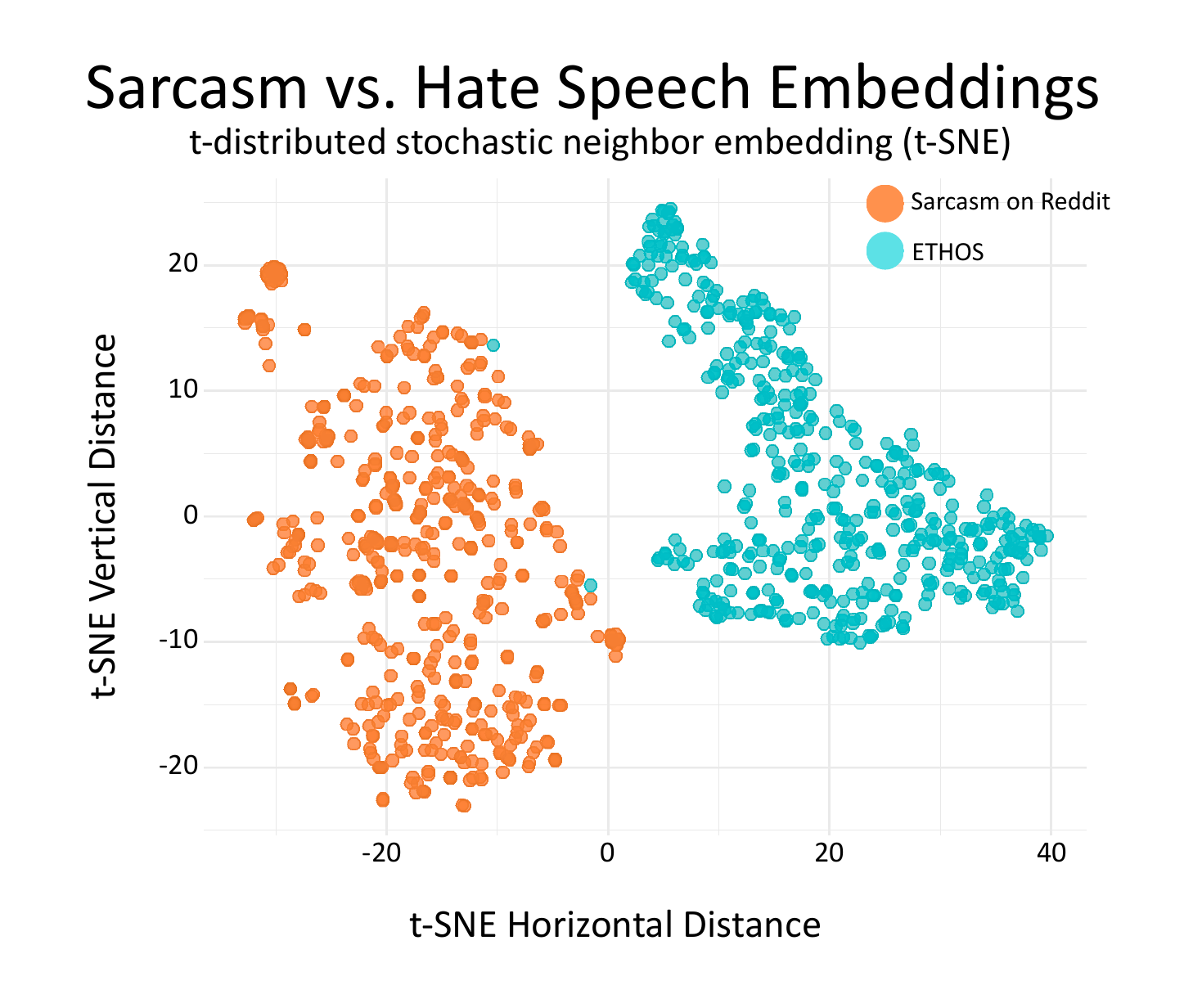}
    \caption{Sarcasm on Reddit vs. ETHOS BERT+BiLSTM embeddings mapped with t-SNE.}
    \label{fig:tSNE_embeddings}
    \vspace{-3mm}
\end{figure}

In order to evaluate whether sarcasm pre-training causes the model to conflate sarcasm with hate speech, we used t-distributed stochastic neighbor embedding (t-SNE) to visualize how well the BERT+BiLSTM separates the two after being trained on both. T-SNE is a non-linear dimensionality reduction technique that preserves local structure in high-dimensional data while enabling two-dimensional visualization~\cite{vandermaaten08a}. As shown in \autoref{fig:tSNE_embeddings}, the model produces clearly separated clusters for sarcasm and hate speech, which indicates that sarcasm pre-training did not lead to systematic confusion between sarcastic and hateful language even in semantically ambiguous cases.

\begin{table}[H]
    \centering
    \captionsetup{justification=centering}
    \small
    \renewcommand{\arraystretch}{1.5} % Row height
    \setlength{\tabcolsep}{8pt} 
    \begin{tabular}{c|c|c|c|c}
        \textbf{Dataset Pair} & \textbf{Top-n} & $\boldsymbol{\mu}$ & \textbf{Min} & \textbf{Max} \\
        \hline
        \shortstack{Sarcasm vs. \\ Corpus} & 1000 & 0.353 & 0.312 & 0.393 \\
        \hline
        \shortstack{Sarcasm vs. \\ ETHOS} & 1000 & 0.353 & 0.183 & 0.576 \\
    \end{tabular}
    \caption{Jaccard similarity metrics with $n$-grams for Sarcasm on Reddit vs. Implicit Hate Corpus and ETHOS.}
    %\vspace{-3mm}
    \label{tab:Jaccard_table}
\end{table}

\begin{table}[H]
    \centering
    \captionsetup{justification=centering}
    \small
    \renewcommand{\arraystretch}{1.5} % Row height
    \begin{tabular}{c|c|c|c|c|c}
        \textbf{Dataset Pair} & \textbf{Sample Size} & $\boldsymbol{\mu}$ & $\boldsymbol{\sigma}$ & \textbf{Min} & \textbf{Max} \\
        \hline
        \shortstack{Sarcasm vs. \\ Corpus} & 1,000 & 0.517 & 0.004 & 0.504 & 0.537 \\
        \hline
        \shortstack{Sarcasm vs. \\ ETHOS} & 500 & 0.611 & 0.010 & 0.588 & 0.666 \\
    \end{tabular}
    \caption{Jensen-Shannon metrics for Sarcasm on Reddit vs. Implicit Hate Corpus and ETHOS.}
    %\vspace{-3mm}
    \label{tab:Jensen_Shannon_table}
\end{table}

\begin{table}[H]
\centering
\captionsetup{justification=centering}
\small
\renewcommand{\arraystretch}{1.5} % Row height
\setlength{\tabcolsep}{8pt} 
\begin{tabular}{c|c|c|c|c}
    \textbf{Similarity Metric} & \textbf{Sample Size} & $\boldsymbol{\mu}$ & \textbf{Min} & \textbf{Max} \\
    \hline
    \shortstack{Jaccard Similarity} & 1000 & 0.369 & 0.332 & 0.407 \\
    \hline
    \shortstack{Jensen-Shannon} & 1000 & 0.496 & 0.471 & 0.546 \\
\end{tabular}
\caption{Sarcasm on Reddit vs. Sarcasm V2 similarity metrics.}
%\vspace{-3mm}
\label{tab:sarcasm_similarity_metrics}
\end{table}

\section{Limitations, Opportunities, \& Future Work}
\begin{figure}[htbp]
    \vspace{-3mm}
    \centering
    \captionsetup{justification=centering}
    \includegraphics[width=\linewidth, keepaspectratio]{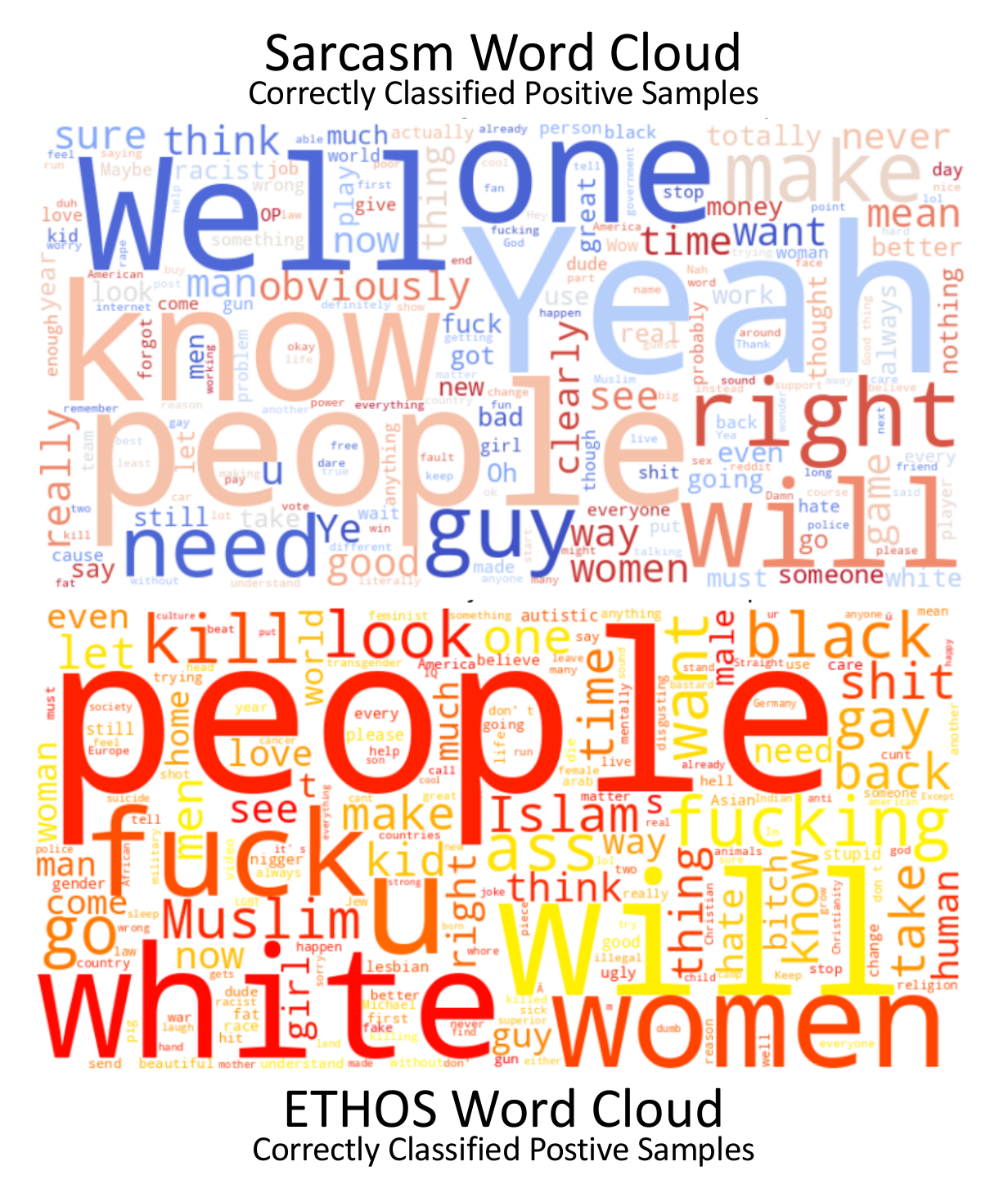}
    \caption{Word clouds of positive samples from ETHOS (hate) and Sarcasm on Reddit (sarcasm).}
    \label{fig:word_clouds}
    \vspace{-3mm}
\end{figure}

\autoref{tab:bert_classifications} shows a selection of the BERT+BiLSTM’s correct and incorrect classifications. These misclassifications indicate that the model relies heavily on charged language to identify hate speech. \autoref{fig:word_clouds} also visualizes frequent terms found in correctly classified ETHOS and Sarcasm on Reddit samples, many of which reference race, gender, religion, and sexual orientation. This overlap suggests a potential risk of the model internalizing biases present in the underlying datasets, whether from prejudice in the source material, subjective or inconsistent labeling, or sampling practices that disproportionately label identity-related comments as hateful. Consequently, the model may learn to associate identity-related terms with hate speech even in neutral or supportive contexts, while overlooking subtler cues. A similar pattern was seen in \autoref{fig:jaccard_venn_diagram}, where over half of the top 1,000 $n$-grams were shared between Sarcasm on Reddit and the Implicit Hate Corpus, and many were derogatory in nature.

More broadly, these issues highlight a fundamental challenge for machine learning in this domain: the tasks we consider lack both abundant data and reliable labels. Standard machine learning assumes that training labels are certain and focuses on reducing uncertainty in predictions for unseen data. In contrast, uncertainty exists even within the training data for hate speech and sarcasm detection. Annotators may disagree on borderline or context-dependent cases, and definitions of abusive speech evolve over time. Sampling choices can further skew the representation toward particular topics or identities, making it difficult to expect high-accuracy predictions from inherently uncertain data. As discussed in \autoref{sec:related_work}, Almohaimeed et al. \cite{almohaimeed2024implicit} demonstrated this challenge when using GPT-4 to assign normal, explicit, and implicit hate labels, then validated by human reviewers. They observed systematic errors, such as mislabeling statements about individuals and conflating opinion with hate speech, resulting in lower performance compared to human annotations. 

These challenges are salient in today’s shifting digital landscape, where evolving platform policies and rising political tensions complicate efforts to define and effectively moderate hate speech. Discussions about free speech and hate speech have become increasingly relevant in the lead-up to the 2024 Presidential Election, as major platforms like Facebook, Instagram, and X relax their policies on hate speech and fact-checking~\cite{ortutay2025metaAP}. Meta’s update to its community standards now allows for ``allegations of mental illness or abnormality'' in discussions around gender and sexual orientation, citing freedom of political and religious discourse as justification \cite{meta2022guidelines}. These shifts create further harm for historically marginalized groups, such as the LGBTQ+ community, which already faces disproportionate targeting online. 

The architecture of an online network shapes how we communicate, what information we encounter, and how we interact with it \cite{HUANG2024106059}. A network's design, whether centralized or decentralized, influences which voices are amplified and how harmful content circulates. However, effective moderation requires time, money, and expertise, which are resources that emerging decentralized platforms lack. At the same time, these limitations create opportunities for platforms like Bluesky to implement stronger moderation systems from the ground up. In 2024, Bluesky received 6.48 million user reports, 17 times the number of reports received the previous year, and its user base expanded from 2.89 million to 25.94 million \cite {2025bluesky}. This growth signals opportunities for automated moderation systems that can better handle covert, abusive, and coded language without reinforcing bias. Hence, future work should prioritize curating more balanced datasets and better representation of slang, euphemisms, and coded language.

\begin{table}[htbp]
    \centering
    \captionsetup{justification=centering}
    \small 
    \renewcommand{\arraystretch}{1.3} 
    \begin{tabular}{|p{3cm}|p{5cm}|}
        \hline
        \textbf{Dataset} & \textbf{Correct / Incorrect Classification}      
        \\ \hline
        \textbf{Sarcasm on Reddit} 
        & \textbf{Correct}: ``Do you really think large corporations would break the rules for money?'' 
        (\textbf{Predicted:} Sarcasm, \textbf{True:} Sarcasm) \\  
        & \textbf{Incorrect}: ``You've yet to meet my university's dining.''
        (\textbf{Predicted:} Sarcasm, \textbf{True:} Non-Sarcasm) \\ \hline
        \textbf{Implicit Hate Corpus} 
        & \textbf{Correct}: ``poll flawed - didn't offer option of enforcement - not just amnesty or deportation.'' 
        (\textbf{Predicted:} Non-Hate, \textbf{True:} Non-Hate) \\  
        & \textbf{Incorrect}: ``All American poor communities black white asian hispanic mexican. I just don't see controversy.''  
        (\textbf{Predicted:} Non-Hate, \textbf{True:} Hate) \\ \hline
        \textbf{ETHOS}              
        & \textbf{Correct}: ``Lgbt now spread in asia. Many anti lgbt in asian worried their culture are ruined cuz this virus.''  
        (\textbf{Predicted:} Hate, \textbf{True:} Hate) \\  
        & \textbf{Incorrect}: ``IM GAY AND FUCKING PROUD FUCK BITCHES LIKE THESE PEOPLE''  
        (\textbf{Predicted:} Hate, \textbf{True:} Non-Hate) \\ \hline
    \end{tabular}
    \caption{BERT+BiLSTM’s correct and incorrect classifications.}
    \label{tab:bert_classifications}
    %\vspace{-6mm}
\end{table}

\section{Conclusion}
This study examined the use of sarcasm as an intermediate task in a transfer learning framework to improve IHS detection. Lexical analyses revealed substantial overlap between sarcasm and IHS, and sarcasm pre-training yielded measurable gains across multiple evaluation metrics, with the largest improvements observed on ETHOS. Although performance on implicit hate was lower than on explicit hate, sarcasm pre-training reduced false positives, limited false negatives, and improved discrimination between hateful and non-hateful content. 

At the same time, our findings highlight broader challenges in this domain. The scarcity of balanced datasets, the subjectivity of annotating labels, and the evolving definitions of abusive speech constrain the reliability of automated detection. Rapidly shifting platform policies further complicate these issues. Future research should pursue technical improvements, such as sarcasm-aware embeddings and multi-task learning, alongside balanced dataset curation to avoid disproportionately labeling identity-related samples as hateful.  As emerging decentralized platforms grow and grapple with moderation, there is an opportunity to design systems that address these challenges proactively. Ultimately, our results demonstrate that leveraging a lexically or semantically related task, such as sarcasm, can help compensate for data scarcity in the target task and fill the gap between overt and covert hate in online spaces, where subtle rhetorical strategies such as sarcasm are increasingly used to mask abusive intent.

\end{document}